\documentclass[10pt,conference]{IEEEtran}
\IEEEoverridecommandlockouts
\usepackage{cite}
\usepackage{amsmath,amssymb,amsfonts}
\usepackage{graphicx}
\usepackage{textcomp}
\usepackage{xcolor}
\usepackage{booktabs}
\usepackage{multirow}
\usepackage{array}
\usepackage{balance}
\usepackage[
    colorlinks=true,
    linkcolor=blue,
    citecolor=blue,
    urlcolor=blue,
    filecolor=blue,
    linktoc=all
]{hyperref}
\usepackage{adjustbox}

\begin{document}

\title{What the Sleeve Feels: Explainable Machine Learning for Textile Pressure-Based Postural Screening}

\author{
\IEEEauthorblockN{Limon Bin Hossain\IEEEauthorrefmark{1}\IEEEauthorrefmark{2} and Md Sadib Rahman Ananta\IEEEauthorrefmark{1}}
\IEEEauthorblockA{\IEEEauthorrefmark{1}Department of Industrial and Production Engineering,\\
Bangladesh University of Engineering and Technology (BUET), Dhaka, 1000, Bangladesh}
\IEEEauthorblockA{\IEEEauthorrefmark{2}Corresponding author: mdlimonbinhossain@gmail.com}
}

\maketitle

\begin{abstract}
Pressure-sensing smart textiles convert body-surface contact into a dense, image-like signal closely tied to posture and movement, making them a promising low-cost route to wearable posture screening. Realizing that promise, however, requires more than classification accuracy: a deployable system must generalize to wearers unseen during training, expose the physical evidence behind its decisions, and tolerate the small donning offsets that occur whenever a garment is removed and re-worn. This paper addresses these three requirements jointly using a knitted piezoresistive sleeve worn on the forearm as a testbed. We regroup fine-grained everyday activities into three coarser screening categories (neutral, potentially undesirable, and functional or transitional), engineer 29 interpretable pressure-distribution features spanning global intensity, spatial center of pressure, quadrant asymmetry, distribution complexity, and short-horizon temporal change, and evaluate under a strict subject-wise split. A tuned XGBoost classifier reaches 0.818 accuracy, 0.788 balanced accuracy, and 0.801 macro F1 on unseen test subjects, with tight frame-level bootstrap 95\% intervals of about $\pm$0.01 and a subject-to-subject standard deviation near 0.06 under leave-one-subject-out cross-validation. A simple 2D-CNN baseline trained on raw frames achieves broadly similar performance, showing that hand-engineered features are not left behind by a learned spatial representation on this task. SHAP-based explanation, a feature-group ablation, per-activity error analysis inside the pooled undesirable class, class-mapping sensitivity, and a simulated donning-rotation stress test together locate what the model relies on, where it degrades, and why, directly targeting the generalization, interpretability, and robustness gaps that determine whether such a system is deployable. 
\end{abstract}


\section{Introduction}

Human activity leaves a continuous imprint on the pressure distribution between the body and its surroundings, whether through contact with a desk, a wall, a chair, or the body itself during folded-arm or self-touch behaviors. Pressure-mapping smart textiles convert this contact into a structured, image-like signal that can be worn, washed, and integrated into clothing, offering a low-cost and unobtrusive alternative to camera-based or inertial motion capture for everyday posture recognition \cite{castano2014,cherenack2012,stoppa2014}. Piezoresistive and capacitive textile structures measure strain and contact pressure directly within the fabric rather than through rigid attached sensors \cite{mattmann2008}, and matrix-style textile arrays have already been demonstrated for seated posture classification in a chair backrest and seat \cite{meyer2010}, for calibrated office-chair posture analysis \cite{xu2013ecushion}, for eating-activity monitoring with a capacitive neckband \cite{cheng2013neckband,chengsj2013}, and for counting gym exercises with a resistive pressure mat \cite{sundholm2014}. These studies confirm that textile pressure matrices carry sufficient spatial information for fine-grained recognition, but nearly all of them instrument a fixed surface, a mat, or a neckband rather than a limb-worn sleeve, and few report performance on wearers unseen during training.

Human activity recognition from body-worn or ambient sensors is a mature area, with comprehensive surveys covering both feature-based and deep-learning pipelines \cite{lara2013,bulling2014,wang2019dl,chen2021survey, straczkiewicz2021}. Within this literature, posture-specific work has applied classical machine learning to pressure and force sensor data for sitting-posture classification \cite{zemp2016} and has produced pressure-map datasets intended for general posture and subject analytics \cite{pouyan2017}. Across this body of work, three requirements that matter for deployment receive limited joint attention: generalization to wearers not seen during training, interpretability of which pressure characteristics drive a given decision, and robustness to the small donning offsets that occur whenever a garment is removed and re-worn. Interpretability matters because a system meant to support ergonomic or rehabilitation feedback cannot act as an opaque classifier, and robustness matters because a model whose decisions rest almost entirely on one spatial axis is fragile in exactly the direction a wearer is most likely to shift the garment.

Ensemble tree methods, particularly random forests \cite{breiman2001} and gradient-boosted trees \cite{chenguestrin2016}, remain strong baselines for tabular sensor features and are the primary and comparison models adopted here. Post-hoc explanation of such models is commonly performed with SHAP (SHapley Additive exPlanations), which attributes a prediction to individual input features through a game-theoretic Shapley value formulation \cite{lundberg2017} and admits an efficient exact implementation for tree ensembles \cite{lundberg2020}, consistent with broader taxonomies of explainable artificial intelligence for building trust in automated decisions \cite{adadi2018,dosilovic2018}.

\begin{figure}[t]
\centering
\includegraphics[width=0.94\columnwidth]{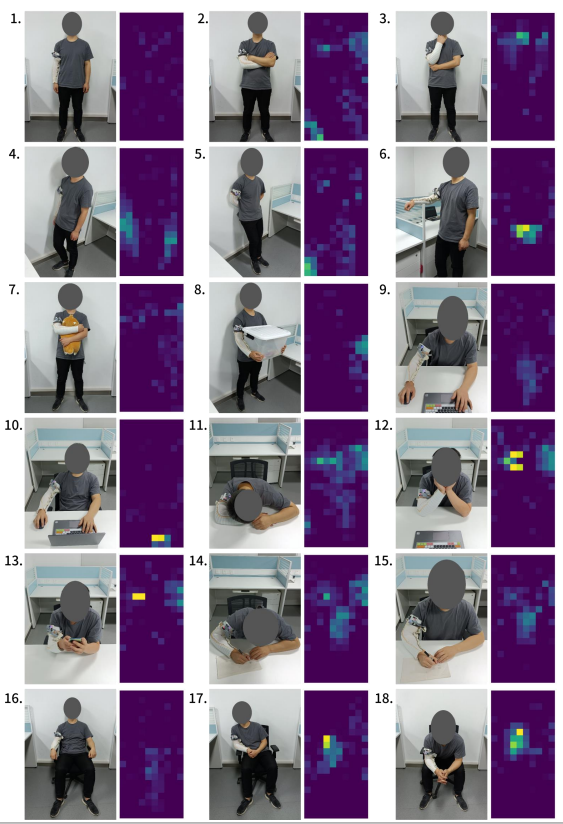}
\caption{Representative photographs and corresponding pressure maps for the 18 everyday activities in the Smart-Sleeve dataset. Image reproduced from the Smart-Sleeve dataset repository \cite{xudataset}.}
\label{fig:activities}
\end{figure}

This paper targets that combined gap: can pressure-distribution features derived from a knitted, matrix-based textile sensor array separate everyday postures into coarse screening categories under a subject-independent protocol, and can the resulting model be probed for the physical evidence behind its decisions, the sources of its errors, and its degradation under realistic donning variation? The sensing platform: a double-sided weft-knitted sleeve whose orthogonally stacked stripes form a 200-point piezoresistive matrix (40 cm $\times$ 18 cm) read out at 12-bit resolution and 50 Hz over Bluetooth. One stripe direction runs along the forearm and the other wraps around its circumference (Fig.~\ref{fig:activities} illustrates the resulting pressure signatures across everyday postures), so the two sensing axes separately encode proximal-distal and angular position, a distinction that becomes central to the explanation results.

The contributions of this paper are as follows.
\begin{itemize}
\item We propose a deterministic, documented mapping from fine-grained everyday activities to three posture-screening categories and engineer 29 interpretable pressure-distribution features, spanning global intensity, activation area, spatial center of pressure, quadrant asymmetry, distribution complexity, and short-horizon temporal change, with all data-driven thresholds estimated from training subjects only.

\item We combine SHAP-based explanation with a feature-group ablation, a per-activity error analysis, and a class-mapping sensitivity study to show which physical pressure characteristics drive each decision and how sensitive the reported performance is to labeling choices.
\item We run a simulated donning-rotation stress test that circularly shifts sensor frames around the limb's circumference, linking the resulting asymmetric degradation directly to the axis identified as most important by the explanation analysis.
\end{itemize}

\section{Methods}

\subsection{Dataset}
The \href{https://github.com/xghgithub/Smart-Sleeve-Dataset}{Smart-Sleeve dataset} contains recordings from 14 healthy, right-handed subjects (3 female) wearing the sleeve on the right arm while performing the 18 everyday activities shown in Fig.~\ref{fig:activities} \cite{xudataset}. Each subject completed 10 rounds of all 18 activities in random order, removing and re-wearing the sleeve between rounds to emulate realistic donning variability, yielding 140 subject-round recordings. Each frame is a 20$\times$10 matrix of 12-bit ADC readings at 50 Hz, and each round's labels file marks the start and end frame index of every activity segment.

\subsection{Screening-Category Mapping}
We define a deterministic, documented mapping from the 18 activity indices to three classes: \textit{neutral or acceptable} (standing, back against the wall, writing with a straight back, sitting with hands on the armrests), \textit{functional or transitional} (folding arms, thinking, side against the wall, arm on the baffle, hugging a doll, carrying a box), and \textit{potentially undesirable} (leaning forward or back at a desk, sleeping on the table, holding one's cheeks, using a mobile phone, writing with a hunched back, sitting leaning to one side, sitting with the arms on the legs). This mapping is a modeling construct based on the semantic content of the original labels, not a validated clinical taxonomy; no activity is labeled as an injury. Two alternative mappings are stress-tested.

\subsection{Feature Engineering}
Let $p_{i,j}$ denote the pressure reading at row $i \in \{1,\ldots,20\}$ and column $j \in \{1,\ldots,10\}$ of a frame, $T=\sum_{i,j} p_{i,j}$ the total pressure, and $r_i=\sum_j p_{i,j}$, $c_j=\sum_i p_{i,j}$ the row and column marginal profiles. Global intensity features are the mean, maximum, minimum, standard deviation, median, 90th and 95th percentiles, total, and range of $\{p_{i,j}\}$, plus two activation-area features recording the fraction of cells above a low and a high threshold, both estimated from training subjects only (Section~\ref{sec:leakage}).

Spatial features describe the center of pressure and its spread along each axis:
\begin{equation}
\bar{y} = \frac{1}{T}\sum_i i\, r_i, \qquad
\bar{x} = \frac{1}{T}\sum_j j\, c_j,
\end{equation}
\begin{equation}
\sigma_y^2 = \frac{1}{T}\sum_i r_i (i-\bar{y})^2, \qquad
\sigma_x^2 = \frac{1}{T}\sum_j c_j (j-\bar{x})^2 .
\end{equation}
Asymmetry features compare pressure across the left-right ($lr$) and up-down ($ud$) halves and the four quadrants, e.g., $lr = (\text{Right}-\text{Left})/T$. Complexity features include the Shannon entropy of the normalized pressure distribution,
\begin{equation}
H = -\sum_{i,j} \frac{p_{i,j}}{T}\log\frac{p_{i,j}}{T}, \quad p_{i,j}>0,
\end{equation}
the concentration of the ten highest-pressure cells relative to $T$, and the ratio of the maximum cell to $T$. Two temporal features, frame-to-frame change in total pressure and frame-to-frame center-of-pressure displacement, are computed within each labeled segment. In total, 29 features are extracted per frame from 84{,}095 labeled frames drawn from 140 subject-round recordings, after sub-sampling every fourth frame (12.5~Hz effective rate) and capping each activity segment at 250 frames to control class imbalance and runtime.

\subsection{Subject-Wise Data Split and Leakage Control}\label{sec:leakage}
Since consecutive frames from the same subject and round are highly correlated, we split at the subject level rather than the frame level: 10 subjects for training (2, 3, 4, 5, 6, 7, 8, 9, 11, 14), 2 for validation (1, 13), and 2 for held-out testing (10, 12), giving 57{,}596 / 13{,}993 / 12{,}506 frames per split. The split is fixed before any threshold estimation or feature computation, and the two activation-percentile thresholds are estimated from training subjects only, so no test-subject frame influences a global preprocessing statistic.

\subsection{Classification Models and Training Protocol}
We compare a multinomial logistic regression baseline with standardized features, a random forest \cite{breiman2001} with 300 trees, and XGBoost \cite{chenguestrin2016} as the primary model, all trained with class-balanced sample weights. XGBoost hyperparameters are selected via a 20-configuration randomized search with 5-fold grouped cross-validation on the training subjects (macro F1 objective, grouped by subject ID), refit with early stopping monitored on the validation subjects, and evaluated exactly once on the test subjects. XGBoost minimizes the regularized additive objective
\begin{equation}
\mathcal{L}^{(t)} = \sum_{k} l\!\left(y_k, \hat{y}_k^{(t-1)}+f_t(x_k)\right) + \Omega(f_t),
\end{equation}
\begin{equation}
\Omega(f) = \gamma T_{\text{leaves}} + \tfrac{1}{2}\lambda \lVert w \rVert^2,
\end{equation}
where $f_t$ is the tree added at boosting round $t$ and $\Omega$ penalizes tree complexity \cite{chenguestrin2016}. The selected configuration used 500 boosted trees at maximum depth 4 and learning rate 0.03, with early stopping selecting the model at iteration 311.

To test whether hand-engineered features are outperformed by a learned spatial representation, we additionally train a small 2D-CNN directly on the raw 20$\times$10 frames (two 3$\times$3 conv blocks with batch normalization, global average pooling, two fully connected layers), using the same subject-wise split, class-frequency-weighted cross-entropy loss, Adam at $10^{-3}$, and early stopping on validation macro F1.

\section{Results and Discussion}
\label{sec:results}

\begin{figure}[t]
\centering
\includegraphics[width=\columnwidth]{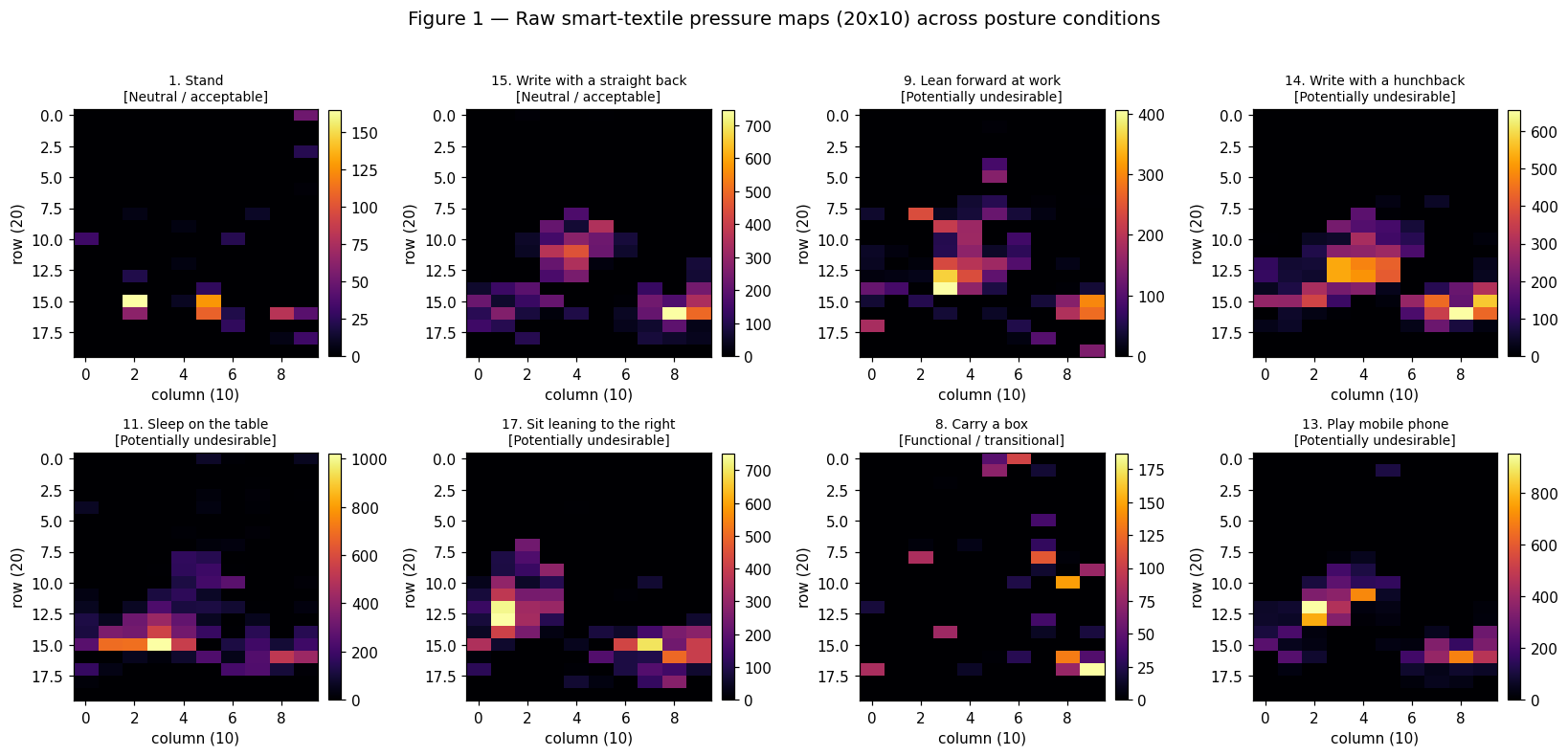}
\caption{Representative raw $20\times10$ pressure frames for eight activities spanning the three screening categories.}
\label{fig:pressuremaps}
\end{figure}

\subsection{Qualitative and Univariate Feature Patterns}
Fig.~\ref{fig:pressuremaps} shows one representative frame per activity across the three screening categories. Neutral postures (standing, writing with a straight back) spread pressure over a diffuse mid-length band with no dominant taxel, whereas undesirable postures (sleeping on the table, sitting leaning to the right, writing with a hunched back, using a mobile phone) concentrate pressure into a tight, high-intensity cluster whose location varies by activity. Carrying a box, the functional example, instead produces several smaller peaks near the sleeve edges, consistent with a two-point grip rather than sustained contact. This qualitative pattern is echoed at the univariate level: total pressure separates the neutral class from the other two, with a markedly lower median and interquartile range, while the left-right balance feature \emph{lr\_diff} takes opposite signs for the functional and undesirable classes, indicating that these categories occupy different halves of the sleeve on average. In contrast, the vertical center of pressure \emph{cop\_y} and the fraction of activated taxels \emph{active\_pct} overlap heavily across all three classes.

\subsection{Feature Table and Class Balance}
The engineered feature table contains 84{,}095 labeled frames and 29 features, with per-subject counts ranging from 4{,}489 (subject 2) to 8{,}911 (subject 13). Class balance is uneven by design: potentially undesirable frames make up about 48\%, functional or transitional about 30\%, and neutral about 22\%, reflecting longer sustained segments for undesirable postures in the source recordings rather than any resampling choice. We therefore report balanced accuracy and macro F1 alongside accuracy throughout.

\subsection{Test-Set Performance and Confidence Intervals}
Table~\ref{tab:models} summarizes the three tabular models on unseen test subjects. XGBoost is the strongest classifier on every metric (XGBoost $>$ random forest $>$ logistic regression). Its accuracy-balanced accuracy gap narrows to 3.0 points, versus 6.2 points for logistic regression, showing it is relatively stronger on the smaller classes, and its ROC AUC of 0.942 exceeds its accuracy of 0.818, indicating well-separated class probabilities that leave room for threshold tuning. A bootstrap of 2{,}000 test-frame resamples gives narrow 95\% intervals, accuracy $[0.812, 0.825]$, balanced accuracy $[0.780, 0.796]$, macro F1 $[0.794, 0.808]$, each about $\pm 0.01$, reflecting frame-level resampling variance rather than subject-to-subject variance, which is addressed by the LOSO study in Section~\ref{sec:loso}.

\begin{table}[t]
\centering
\caption{Test-set performance on unseen subjects (subjects 10, 12; 12{,}506 frames)}
\label{tab:models}
\begin{tabular}{lcccc}
\toprule
Model & Acc. & Bal.\ Acc. & Macro F1 & ROC AUC \\
\midrule
Logistic Regression & 0.730 & 0.668 & 0.681 & 0.873 \\
Random Forest       & 0.793 & 0.737 & 0.762 & 0.933 \\
XGBoost             & \textbf{0.818} & \textbf{0.788} & \textbf{0.801} & \textbf{0.942} \\
TinyCNN (raw frames) & 0.814 & 0.779 & 0.795 & --- \\
\bottomrule
\end{tabular}
\end{table}

\subsection{Per-Class Confusion and Sub-Cluster Behavior}
\label{sec:subcluster}
\begin{figure}[t]
\centering
\includegraphics[width=0.85\columnwidth]{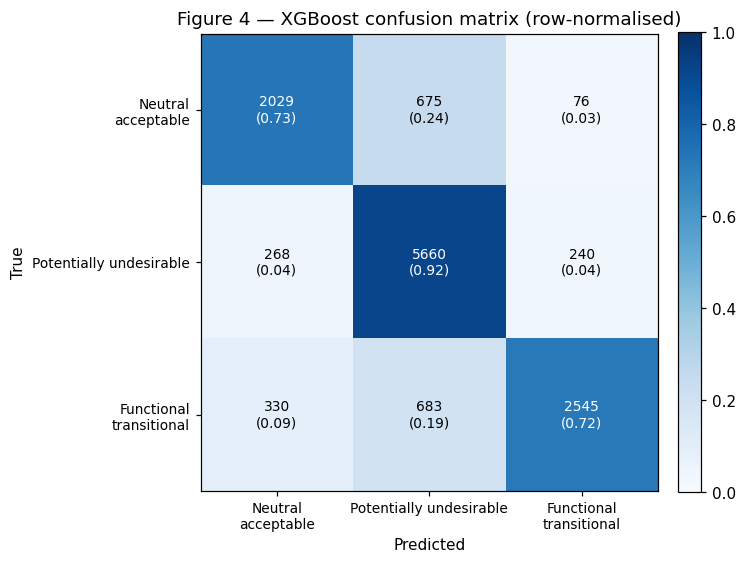}
\caption{Row-normalized confusion matrix for XGBoost on the held-out test subjects.}
\label{fig:confusionviz}
\end{figure}

The confusion matrix ( Fig.~\ref{fig:confusionviz}) shows that the undesirable class acts as an attractor for ambiguous frames: its predicted column holds the highest diagonal value (0.92) and the two largest off-diagonal entries in the matrix (0.24 from neutral, 0.19 from functional), while direct neutral-functional confusion stays at or below 0.09. Table~\ref{tab:persubj} shows this pattern varies by subject: subject 10 tracks the aggregate closely, while subject 12's balanced accuracy and macro F1 sit about six points lower, an early hint of the subject variance quantified next by LOSO.

\begin{table}[t]
\centering
\caption{Per-test-subject XGBoost metrics on the two held-out subjects. Subject 10's numbers are close to the aggregate test figures, while subject 12's balanced accuracy and macro F1 sit below them, foreshadowing the subject-level spread quantified by leave-one-subject-out cross-validation.}
\label{tab:persubj}
\setlength{\tabcolsep}{4pt}
\begin{tabular}{ccccc}
\toprule
Subj. & Frames & Acc. & Bal.\ Acc. & Macro F1 \\
\midrule
10 & 6{,}752 & 0.832 & 0.822 & 0.828 \\
12 & 5{,}754 & 0.802 & 0.744 & 0.763 \\
\bottomrule
\end{tabular}
\end{table}

Because the undesirable class pools eight source activities, we further ask how uniformly the model recognizes them (Table~\ref{tab:subcluster}). Three activities reach recall 1.00 (sleeping on the table, holding cheeks, sitting leaning to the right) and two more exceed 0.94 (writing with a hunched back, using a mobile phone), while ``sit with arms on the legs'' (0.83) and ``lean back at work'' (0.74) are hardest, with the latter's misclassified frames routed mostly to neutral (17.7\%). The easiest activities are exactly those with a dense, concentrated contact signature, while the hardest present a more diffuse pattern; the pooled undesirable label therefore hides real heterogeneity that a single recall figure would not reveal.

\begin{table}[t]
\centering
\caption{Per-activity behavior of XGBoost inside the pooled potentially undesirable class on the test subjects. }
\label{tab:subcluster}
\setlength{\tabcolsep}{2.5pt}
\begin{tabular}{lccccc}
\toprule
Activity & Frames & Recall & \% N & \% U & \% F \\
\midrule
Lean back at work         & 661 & 0.735 & 17.7 & 73.5 &  8.8 \\
Sit with arms on the legs & 844 & 0.829 &  0.0 & 82.9 & 17.1 \\
Lean forward at work      & 710 & 0.856 & 14.4 & 85.6 &  0.0 \\
Play mobile phone         & 749 & 0.944 &  0.5 & 94.4 &  5.1 \\
Write with a hunchback    & 834 & 0.946 &  5.4 & 94.6 &  0.0 \\
Sleep on the table        & 723 & 1.000 &  0.0 & 100.0 &  0.0 \\
Hold cheeks               & 814 & 1.000 &  0.0 & 100.0 &  0.0 \\
Sit leaning to the right  & 833 & 1.000 &  0.0 & 100.0 &  0.0 \\
\bottomrule
\end{tabular}
\end{table}

\subsection{Leave-One-Subject-Out Cross-Validation}
\label{sec:loso}
\begin{figure}[t]
\centering
\includegraphics[width=\columnwidth]{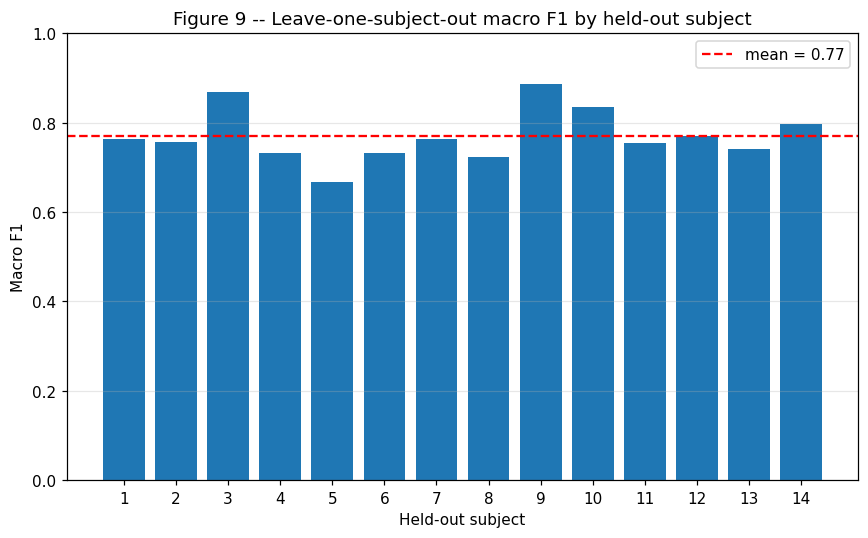}
\caption{Leave-one-subject-out macro F1 for XGBoost, one bar per held-out subject; red dashed line marks the mean.}
\label{fig:loso}
\end{figure}

Retraining XGBoost with the selected hyperparameters and holding each of the 14 subjects out in turn (Fig.~\ref{fig:loso}) gives a mean macro F1 of 0.770 (SD 0.059, range 0.666 to 0.887) and similarly distributed accuracy (mean 0.785, SD 0.061, range 0.660 to 0.894). The two subjects held out for the main test split (10 and 12) land at 0.834 and 0.769 macro F1, both above the mean, so the 0.801 test macro F1 reported earlier sits on the upper side of the subject distribution. The $\pm 0.06$ subject-level spread is roughly six times wider than the frame-level bootstrap interval, confirming that subject variance, not test-set noise, is the dominant source of residual uncertainty and the figure to consult when considering deployment to new wearers.

\subsection{Feature-Group Ablation}
\label{sec:ablation}
Retraining XGBoost on each of the six engineered feature families in isolation (Table~\ref{tab:ablation}) reaches 0.80 macro F1 for the full set, led by spatial features alone at 0.64, followed by global intensity (0.57), asymmetry (0.54), complexity (0.50), temporal (0.45), and activation area (0.40, the weakest). The spatial family alone recovers most of the gap from the majority-class baseline, showing that where contact occurs on the sleeve matters more than how much of the sleeve is touched. Still, the 16-point gap between spatial alone (0.64) and the full model (0.80) shows that no single family is sufficient: XGBoost benefits from combining spatial cues with intensity, asymmetry, complexity, and temporal change, consistent with the SHAP ranking below, which shows the top features as important but not solely adequate.

\begin{table}[t]
\centering
\caption{Feature-group ablation on the held-out test subjects.}
\label{tab:ablation}
\setlength{\tabcolsep}{4pt}
\begin{tabular}{lcccc}
\toprule
Feature group & \# feats & Acc. & Bal.\ Acc. & Macro F1 \\
\midrule
All              & 29 & 0.818 & 0.788 & \textbf{0.801} \\
Spatial          & 6  & 0.688 & 0.639 & 0.643 \\
Global intensity & 9  & 0.629 & 0.569 & 0.574 \\
Asymmetry        & 6  & 0.582 & 0.539 & 0.539 \\
Complexity       & 3  & 0.505 & 0.511 & 0.495 \\
Temporal         & 2  & 0.486 & 0.447 & 0.449 \\
Activation area  & 3  & 0.460 & 0.397 & 0.405 \\
\bottomrule
\end{tabular}
\end{table}

\subsection{Sensitivity to the Class-Mapping Scheme}
\label{sec:mappingsens}
Because the three-class mapping is a design choice rather than a property of the dataset, we test two alternatives without changing anything else. Mapping A swaps ``arm on the baffle'' into undesirable and ``sit with hands on the armrests'' into functional. Mapping B moves ``sitting leaning to the right'' and ``sit with arms on the legs'' out of undesirable and into functional. Both give macro F1 close to the original (0.810 and 0.815 versus 0.801; Table~\ref{tab:mappingsens}), with Mapping B also improving balanced accuracy (0.815 versus 0.788) as expected when a heterogeneous class is trimmed. The reported performance is therefore not an artifact of one arbitrary mapping choice.

\begin{table}[t]
\centering
\caption{Test-set XGBoost performance under three three-class mappings evaluated on the same held-out subjects.}
\label{tab:mappingsens}
\setlength{\tabcolsep}{4pt}
\begin{tabular}{lccc}
\toprule
Mapping & Acc. & Bal.\ Acc. & Macro F1 \\
\midrule
Original                       & 0.818 & 0.788 & 0.801 \\
Alt. A (borderline swap)       & 0.841 & 0.790 & 0.810 \\
Alt. B (stricter undesirable)  & 0.827 & 0.815 & 0.815 \\
\bottomrule
\end{tabular}
\end{table}

\subsection{Explainability via SHAP}
\label{sec:shap}
\begin{figure}[t]
\centering
\includegraphics[width=\columnwidth]{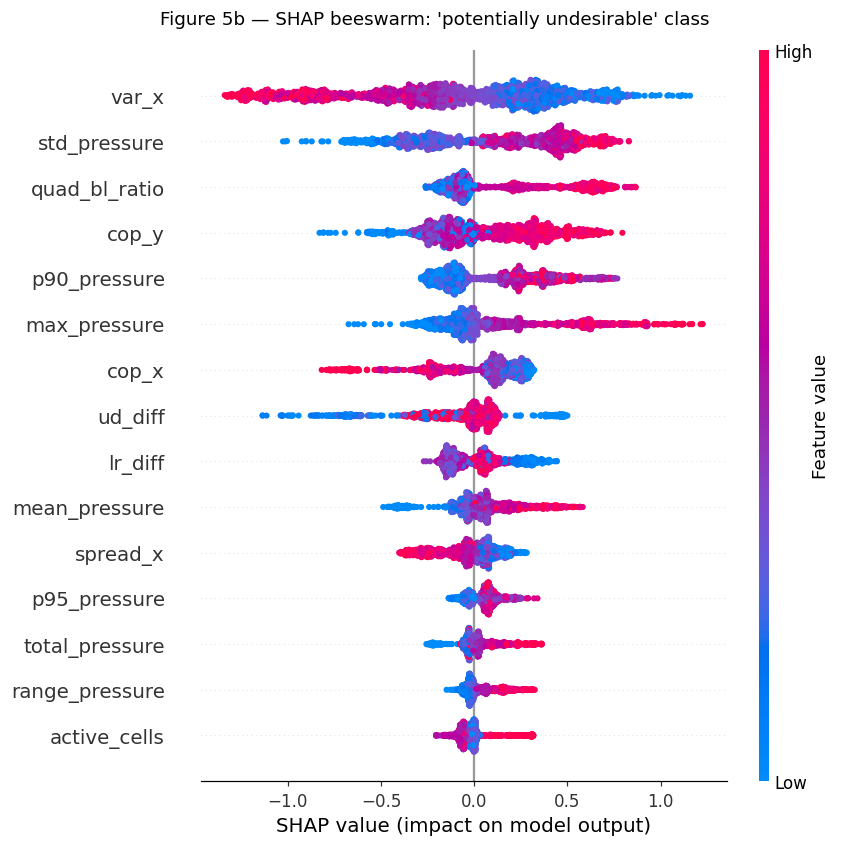}
\caption{SHAP beeswarm plot for the potentially undesirable class. Each point is one test-set frame; horizontal position is the feature's SHAP value and color encodes the feature's own numerical value from low (blue) to high (red).}
\label{fig:beeswarm}
\end{figure}

The two most influential features, \emph{cop\_x} and \emph{var\_x}, describe the position and spread of pressure along the 10-taxel circumferential axis and outweigh \emph{std\_pressure} and \emph{cop\_y}, which describe intensity variability and lengthwise position. For this sleeve geometry, where pressure wraps around the forearm therefore matters more to the model than where along its length contact occurs, plausibly because everyday postures differ more in which side of the forearm makes contact than in how far up or down it extends. The beeswarm plot (Fig.~\ref{fig:beeswarm}) shows clean color separation for \emph{var\_x} and \emph{std\_pressure} (high circumferential spread pushes away from undesirable; high pressure variability pushes toward it), while lower-ranked features such as \emph{lr\_diff} and \emph{ud\_diff} overlap on both sides of zero, indicating smaller, more context-dependent contributions. The features driving individual frame decisions match this global ranking, so the pattern is not an artifact of averaging over dissimilar frames, though exact magnitudes and occasional signs still vary frame to frame, reflecting frame-specific reweighting of the same handful of physical properties.

\subsection{Robustness to Simulated Donning Rotation}
\label{sec:rotation}
Evaluating the frozen XGBoost model on test frames whose 10-column axis is circularly rotated by $\pm 1$, $\pm 2$, and $\pm 3$ taxels (Table~\ref{tab:rotation}) shows an asymmetric, non-negligible degradation: macro F1 falls from 0.80 at shift 0 to 0.79, 0.73, and 0.62 for positive shifts of one, two, and three taxels, and to 0.72, 0.52, and 0.47 for the corresponding negative shifts. The negative direction consistently hurts more, consistent with \emph{cop\_x} being the top SHAP feature: rotating the pressure image around the circumference directly perturbs the axis the model relies on most. A two-taxel donning offset alone would cost about seven macro-F1 points in the positive direction and nearly 28 in the negative direction, a real deployment concern that would require a wearing-orientation marker, an alignment step at donning, or rotation-augmented training.

\begin{table}[t]
\centering
\caption{Robustness of the frozen XGBoost model to simulated circumferential donning rotation.}
\label{tab:rotation}
\setlength{\tabcolsep}{5pt}
\begin{tabular}{cccc}
\toprule
Shift (taxels) & Acc. & Bal.\ Acc. & Macro F1 \\
\midrule
$-3$ & 0.533 & 0.489 & 0.471 \\
$-2$ & 0.614 & 0.541 & 0.522 \\
$-1$ & 0.761 & 0.706 & 0.721 \\
 $0$ & \textbf{0.818} & \textbf{0.788} & \textbf{0.801} \\
$+1$ & 0.807 & 0.775 & 0.787 \\
$+2$ & 0.749 & 0.719 & 0.728 \\
$+3$ & 0.634 & 0.606 & 0.618 \\
\bottomrule
\end{tabular}
\end{table}
\section{Conclusion}
This paper presented an explainable re-analysis of the Smart-Sleeve pressure-mapping dataset for coarse postural screening. Using 29 interpretable features and a strict subject-independent protocol, XGBoost reached 0.818 accuracy, 0.788 balanced accuracy, and 0.801 macro F1 on unseen subjects, with a bootstrap 95\% interval of about $\pm 0.01$ on those metrics and a LOSO subject-level spread of about $\pm 0.06$. A simple raw-frame 2D-CNN reached essentially the same performance, so at this dataset scale the interpretable tabular pipeline is not left behind by a learned spatial representation. SHAP-based explanation, together with a feature-group ablation and a sub-cluster analysis, showed that circumferential spatial features contribute more than raw pressure magnitude and that the pooled undesirable class hides real heterogeneity across its source activities. A donning-rotation stress test quantified how much orientation matters for this feature set. These findings support the feasibility of coupling textile pressure sensing with explainable machine learning for low-cost, interpretable posture screening, while making the orientation dependence and the mapping-driven confusion structure explicit rather than hidden.

\section{Limitations and Future Work}
The present study has several limitations that should guide future work. The screening categories are derived heuristically from activity semantics rather than from measured biomechanical or clinical outcomes, and the dataset provides no injury or discomfort labels, so no claim of injury prediction or risk diagnosis is made or implied. The sample (14 right-handed subjects, sleeve on the right arm) also limits demographic and anatomical diversity. Future work should therefore evaluate the pipeline on a larger, more diverse subject pool; extend the rotation study to the lengthwise axis and combine rotation with training-time augmentation for orientation-tolerant variants; explore sequence models that exploit longer temporal context across activity transitions rather than within-segment features alone; benchmark directly against recent pressure-sensing screening pipelines under identical subject-independent protocols; and, where ethically and clinically appropriate, validate the screening categories against expert ergonomic assessment rather than activity labels alone.

\balance
\bibliographystyle{IEEEtran}
\bibliography{refs}

@article{castano2014,
  author  = {Castano, L. M. and Flatau, A. B.},
  title   = {Smart fabric sensors and e-textile technologies: a review},
  journal = {Smart Materials and Structures},
  volume  = {23},
  number  = {5},
  pages   = {053001},
  year    = {2014}
}

@article{cherenack2012,
  author  = {Cherenack, K. and van Pieterson, L.},
  title   = {Smart textiles: Challenges and opportunities},
  journal = {Journal of Applied Physics},
  volume  = {112},
  number  = {9},
  pages   = {091301},
  year    = {2012}
}

@article{stoppa2014,
  author  = {Stoppa, M. and Chiolerio, A.},
  title   = {Wearable electronics and smart textiles: A critical review},
  journal = {Sensors},
  volume  = {14},
  number  = {7},
  pages   = {11957--11992},
  year    = {2014}
}

@article{mattmann2008,
  author  = {Mattmann, C. and Clemens, F. and Tr{\"o}ster, G.},
  title   = {Sensor for measuring strain in textile},
  journal = {Sensors},
  volume  = {8},
  number  = {6},
  pages   = {3719--3732},
  year    = {2008}
}

@article{meyer2010,
  author  = {Meyer, J. and Arnrich, B. and Schumm, J. and Tr{\"o}ster, G.},
  title   = {Design and modeling of a textile pressure sensor for sitting posture classification},
  journal = {IEEE Sensors Journal},
  volume  = {10},
  number  = {8},
  pages   = {1391--1398},
  year    = {2010}
}

@article{xu2013ecushion,
  author  = {Xu, W. and Huang, M.-C. and Amini, N. and He, L. and Sarrafzadeh, M.},
  title   = {{eCushion}: A textile pressure sensor array design and calibration for sitting posture analysis},
  journal = {IEEE Sensors Journal},
  volume  = {13},
  number  = {10},
  pages   = {3926--3934},
  year    = {2013}
}

@inproceedings{cheng2013neckband,
  author    = {Cheng, J. and Zhou, B. and Kunze, K. and Rheinl{\"a}nder, C. C. and Wille, S. and Wehn, N. and Weppner, J. and Lukowicz, P.},
  title     = {Activity recognition and nutrition monitoring in every day situations with a textile capacitive neckband},
  booktitle = {Proc. ACM Conf. Pervasive and Ubiquitous Computing Adjunct Publication (UbiComp)},
  pages     = {155--158},
  year      = {2013}
}

@article{chengsj2013,
  author  = {Cheng, J. and Amft, O. and Bahle, G. and Lukowicz, P.},
  title   = {Designing sensitive wearable capacitive sensors for activity recognition},
  journal = {IEEE Sensors Journal},
  volume  = {13},
  number  = {10},
  pages   = {3935--3947},
  year    = {2013}
}

@inproceedings{sundholm2014,
  author    = {Sundholm, M. and Cheng, J. and Zhou, B. and Sethi, A. and Lukowicz, P.},
  title     = {Smart-mat: Recognizing and counting gym exercises with low-cost resistive pressure sensing matrix},
  booktitle = {Proc. ACM Int. Joint Conf. Pervasive and Ubiquitous Computing (UbiComp)},
  pages     = {373--382},
  year      = {2014}
}

@article{lara2013,
  author  = {Lara, O. D. and Labrador, M. A.},
  title   = {A survey on human activity recognition using wearable sensors},
  journal = {IEEE Communications Surveys \& Tutorials},
  volume  = {15},
  number  = {3},
  pages   = {1192--1209},
  year    = {2013}
}

@article{bulling2014,
  author  = {Bulling, A. and Blanke, U. and Schiele, B.},
  title   = {A tutorial on human activity recognition using body-worn inertial sensors},
  journal = {ACM Computing Surveys},
  volume  = {46},
  number  = {3},
  pages   = {33},
  year    = {2014}
}

@article{wang2019dl,
  author  = {Wang, J. and Chen, Y. and Hao, S. and Peng, X. and Hu, L.},
  title   = {Deep learning for sensor-based activity recognition: A survey},
  journal = {Pattern Recognition Letters},
  volume  = {119},
  pages   = {3--11},
  year    = {2019}
}

@article{chen2021survey,
  author  = {Chen, K. and Zhang, D. and Yao, L. and Guo, B. and Yu, Z. and Liu, Y.},
  title   = {Deep learning for sensor-based human activity recognition: Overview, challenges, and opportunities},
  journal = {ACM Computing Surveys},
  volume  = {54},
  number  = {4},
  pages   = {77},
  year    = {2021}
}

@article{straczkiewicz2021,
  author  = {Straczkiewicz, M. and James, P. and Onnela, J.-P.},
  title   = {A systematic review of smartphone-based human activity recognition methods for health research},
  journal = {npj Digital Medicine},
  volume  = {4},
  pages   = {148},
  year    = {2021}
}

@article{zemp2016,
  author  = {Zemp, R. and Tanadini, M. and Pl{\"u}ss, S. and Schn{\"u}riger, K. and Singh, N. B. and Taylor, W. R. and Lorenzetti, S.},
  title   = {Application of machine learning approaches for classifying sitting posture based on force and acceleration sensors},
  journal = {BioMed Research International},
  volume  = {2016},
  pages   = {5978489},
  year    = {2016}
}

@inproceedings{pouyan2017,
  author    = {Pouyan, M. B. and Birjandtalab, J. and Heydarzadeh, M. and Nourani, M. and Ostadabbas, S.},
  title     = {A pressure map dataset for posture and subject analytics},
  booktitle = {Proc. IEEE EMBS Int. Conf. Biomedical and Health Informatics (BHI)},
  pages     = {65--68},
  year      = {2017}
}

@article{breiman2001,
  author  = {Breiman, L.},
  title   = {Random forests},
  journal = {Machine Learning},
  volume  = {45},
  number  = {1},
  pages   = {5--32},
  year    = {2001}
}

@inproceedings{chenguestrin2016,
  author    = {Chen, T. and Guestrin, C.},
  title     = {{XGBoost}: A scalable tree boosting system},
  booktitle = {Proc. 22nd ACM SIGKDD Int. Conf. Knowledge Discovery and Data Mining (KDD)},
  pages     = {785--794},
  year      = {2016}
}

@inproceedings{lundberg2017,
  author    = {Lundberg, S. M. and Lee, S.-I.},
  title     = {A unified approach to interpreting model predictions},
  booktitle = {Advances in Neural Information Processing Systems (NeurIPS)},
  pages     = {4765--4774},
  year      = {2017}
}

@article{lundberg2020,
  author  = {Lundberg, S. M. and Erion, G. and Chen, H. and DeGrave, A. and Prutkin, J. M. and Nair, B. and Katz, R. and Himmelfarb, J. and Bansal, N. and Lee, S.-I.},
  title   = {From local explanations to global understanding with explainable {AI} for trees},
  journal = {Nature Machine Intelligence},
  volume  = {2},
  number  = {1},
  pages   = {56--67},
  year    = {2020}
}

@article{adadi2018,
  author  = {Adadi, A. and Berrada, M.},
  title   = {Peeking inside the black-box: A survey on explainable artificial intelligence ({XAI})},
  journal = {IEEE Access},
  volume  = {6},
  pages   = {52138--52160},
  year    = {2018}
}

@inproceedings{dosilovic2018,
  author    = {Do{\v{s}}ilovi{\'c}, F. K. and Br{\v{c}}i{\'c}, M. and Hlupi{\'c}, N.},
  title     = {Explainable artificial intelligence: A survey},
  booktitle = {Proc. 41st Int. Convention on Information and Communication Technology, Electronics and Microelectronics (MIPRO)},
  pages     = {0210--0215},
  year      = {2018}
}

@misc{xudataset,
  author       = {Xu, G. and Deng, W. and Cheng, J.},
  title        = {{Smart-Sleeve} dataset based on pressure mapping smart textile},
  howpublished = {GitHub repository, University of Science and Technology of China},
  note         = {Available: \url{https://github.com/xghgithub/Smart-Sleeve-Dataset}},
  year         = {2023}
}

\end{document}